\documentclass{article}
\usepackage{authblk}
\usepackage{hyperref} % Dodajemy pakiet hyperref

\usepackage{graphicx} % Required for inserting images
\usepackage{subfig}
\usepackage{algorithmic}
\usepackage{algorithm}
\usepackage{listings}
\usepackage[cmex10]{amsmath}
\usepackage{url}
\usepackage{multirow}
\title{Three-Dimensional Path Planning: Navigating through Rough Mereology}

\author[1]{Autor Aleksandra Szpakowska\thanks{\href{mailto:ola.szpakowska@matman.uwm.edu.pl}{ola.szpakowska@matman.uwm.edu.pl}}}
\author[1]{Piotr Artiemjew\thanks{\href{mailto:artem@matman.uwm.edu.pl}{artem@matman.uwm.edu.pl}}}
\affil[1]{University of Warmia and Mazury in Olsztyn\\
ul. S{\l}oneczna 54, 10-710 Olsztyn, Poland\\}

\begin{document}
% \institute{University of Warmia and Mazury in Olsztyn\\
% ul. S{\l}oneczna 54, 10-710 Olsztyn, Poland\\
% \email{ola.szpakowska@matman.uwm.edu.pl, artem@matman.uwm.edu.pl}}

\maketitle

\begin{abstract}In this paper, we present an innovative technique for the path planning of flying robots in a 3D environment in Rough Mereology terms. The main goal was to construct the algorithm that would generate the mereological potential fields in 3-dimensional space. To avoid falling into the local minimum, we assist with a weighted Euclidean distance. Moreover, a searching path from the start point to the target, with respect to avoiding the obstacles was applied. The environment was created by connecting two cameras working in real-time. To determine the gate and elements of the world inside the map was responsible the Python Library OpenCV \cite{OpenCV} which recognized shapes and colors. The main purpose of this paper is to apply the given results to drones.

\end{abstract}
% \newpage
\section{Introduction}

In the field of mobile robotics, tasks such as path planning necessitate a comprehensive representation of the robot's surroundings, including obstacles, landmarks and other robots, coupled with an efficient reasoning framework. Mobile robotics draws upon a wealth of concepts and techniques from Computer Science and Artificial Intelligence, utilizing graph-based methods and algorithms like A* for search or planning efforts, graph mappings for constructing maps, use of computer vision for map navigation among others. In our research, we are continuing to develop the innovative concept of path planning by means of post-map expansion via a mereological potential field algorithm, which was originally introduced in the works of Polkowski and Osmialowski \cite{polkowski2008}. The theoretical foundations were introduced by Polkowski in the work \cite{Polkowski 1996}.  This paper focuses on the application of the main principles of rough mereology \cite{Polkowski 1996} to the three-dimensional environment. The study extends the work of \cite{SzpakowskaArtiemjew 2023}, where approximate fields of mereological potential were employed to explore the three-dimensional map, respecting the distribution of obstacles in the environment. Additionally, an algorithm for searching the created potential fields was implemented, where the selection is based on weighted distance calculation to the goal, followed by selecting the field with the minimum value of the computed distance. To improve the results, a path optimization function was initiated, along with subsequent path smoothing. The environment from which the map was created was developed using \cite{OpenCV}. The map - gate was adjusted to real dimensions, actual distances between markers were measured and transferred to the virtual world, and the coordinates of map points were normalized based on real-world distances. The goal of the paper is to demonstrate a ready-to-use tool for application on a devices, such as a drones.

\subsection{Use of Rough Mereology in the Control Environment of Intelligent Agents}\label{methodology-sect}

This section explores the application of rough mereology for generating potential fields. The application of rough mereology introduces the notion of rough inclusion, symbolized as $\mu (x,y,r)$. This concept asserts that $x$ is partially included in $y$ with a minimum degree of $r$. Focusing on spatial entities, rough inclusion is defined as $\mu (X,Y,r)$ if and only if $\frac{|X \cap Y|}{|X|} \geq r$, where $X$ and $Y$ denote n-dimensional solids, and $|X|$ represents the n-volume of $X$. This study explores a planar case of an autonomous mobile robot moving within a three-dimensional space. Here, the spatial entities $X,Y$ are considered as conceptual regions, with $|X|$ indicating the area of $X$. The rough inclusion $\mu (X,Y,r)$ plays a role in forming the rough mereological potential field. \newline The constituents of this field are square in shape, and their relative proximity (distance in our rough mereological universe) is quantified as:
$$K(X,Y)=min\{max_r\mu(X,Y,r),max_s \mu(Y,X,s)\}.$$
An elaborate discussion on the construction of this field is found in Sect. \ref{methodology}. 

% The path of the robot through the field to its goal is guided by waypoints, which are established inductively: the next waypoint is determined as the centroid of the combined field squares adjacent to the square containing the current waypoint, in accordance to the distance $K(X,Y)$.

The immediate importance of using rough mereology in creating mereological potential fields is to provide a robust framework for representing, analyzing, and navigating spatial environments, especially in scenarios characterized by uncertainty and complexity. 
Due to the limited size of this paper, for a detailed study of the properties that provide the theoretical foundation for our work, we refer the reader to Section 2.3 of paper \cite{Osmialowski2008}.

% \newline
The subsequent document sections are organized as follows. Section \ref{methodology} introduces the methodology for route planning that employs a rough mereological potential field. In Section \ref{experiments}, we detail the experimental setup used. Conclusively, in Section \ref{conclusions}, we offer a summary of our study.

\section{Methodology}\label{methodology}

In this section, we will focus on the different techniques used to build a target robot guidance system using rough mereological potential fields in a 3D environment. In the following subsections, we will explain the working principle of the algorithm responsible for generating the mereological potential field. Importantly, the initialization value is the target coordinate. Such a declaration contributes to a large accumulation of potential fields relative to the target. In the first iterations of the algorithm, the distance is the smallest (its value increases with subsequent iterations of the algorithm), which results in the accumulation of potential fields around the value that initializes the algorithm. The idea is that the generated potential fields attract our starting point to the goal.

\begin{figure}
    \centering
        \caption{The visualization of an idea of neighbors creating in 3-dimensional space. The N, W, S, E, NE, SE, SW, and NW represent the world direction. Values x, y, z determines 3 dmiensions.}
    \includegraphics[scale =0.2]{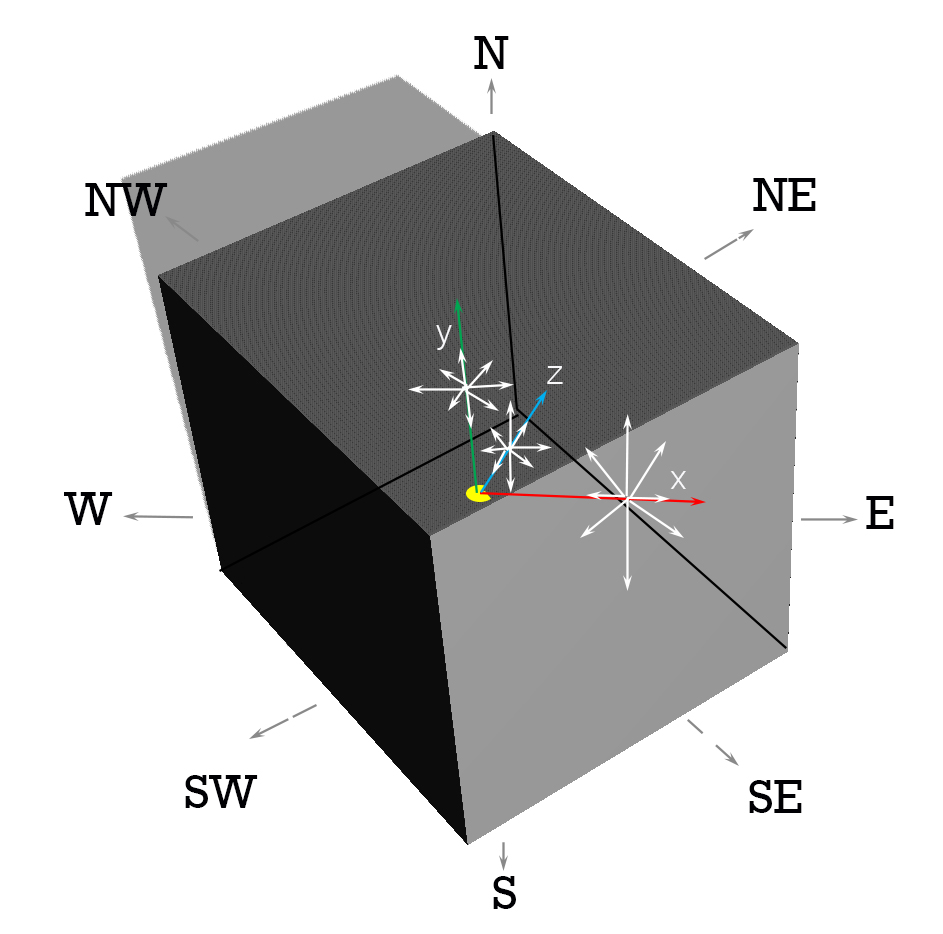}

    \label{directions}
\end{figure}

\subsection{Adapted Square fill algorithm in 3D environment} \label{sqr_alg}
\begin{figure}
\begin{center}
  \caption{The middle point inside the cube determines the specific goal point. Values x,y, and z determine the dimension, +1/-1 is a sample value of the declared distance. In that case, we received 24 new neighbors, which are generated in each iteration of an algorithm. What is important is that the central point determines the goal point only in the first iteration of an algorithm, after that the central point turns to each neighbor created in the previous iteration.}\label{vectors}
  \includegraphics[scale=0.6]{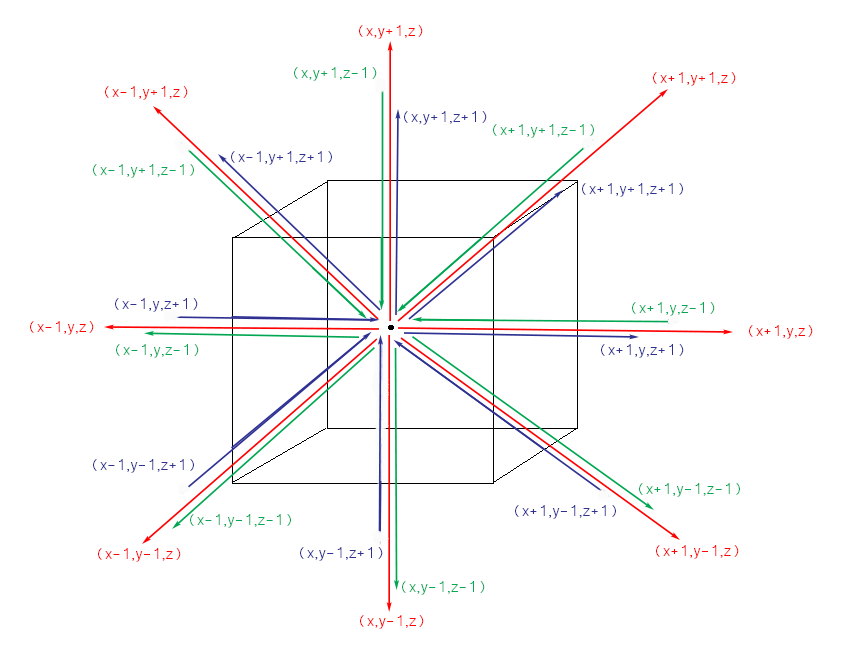}
\end{center}
\end{figure}

In this chapter, we are going to show our conception of a 3-dimensional potential field algorithm introduced by Ośmiałowski \cite{Osmialowski 2011}.   The mentioned Squared Fill algorithm method was already modified and later presented in Polkowski \cite{Polkowski 2018}, Żmudziński and Artiemjew \cite{Zmudzinski Artiemjew 2017} and Szpakowska and Artiemjew \cite{SzpakowskaArtiemjew 2023}. Below we present our approach to generating potential fields in the 3-dimensional environment- Fig. \ref{directions}, 
Fig.\ref{vectors}.

\begin{figure}[ht]

\begin{center}
  \caption{The Rough Mereological Potential Field algorithm constructed in 3-dimensional environment. The below case represents the goal point located at the center of a cube without any obstacles. The bigger cube represents the gate for a robot. }\label{square_algorithm}
  \includegraphics[scale=0.5]{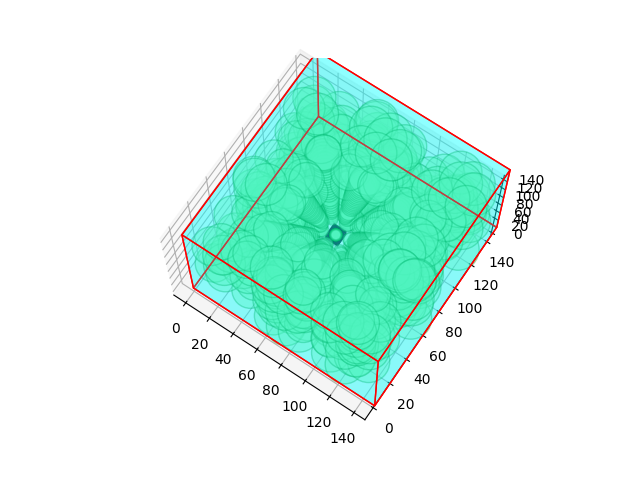}
\end{center}
\end{figure}

\begin{figure}
\begin{center}
  \caption{The Rough Mereological Potential Field algorithm constructed in 3-dimensional environment. The below case represents the goal point located at the center of a cube with only one obstacle. The bigger cube represents the gate for a robot. }\label{square_algorithm2}
  \includegraphics[scale=0.35]{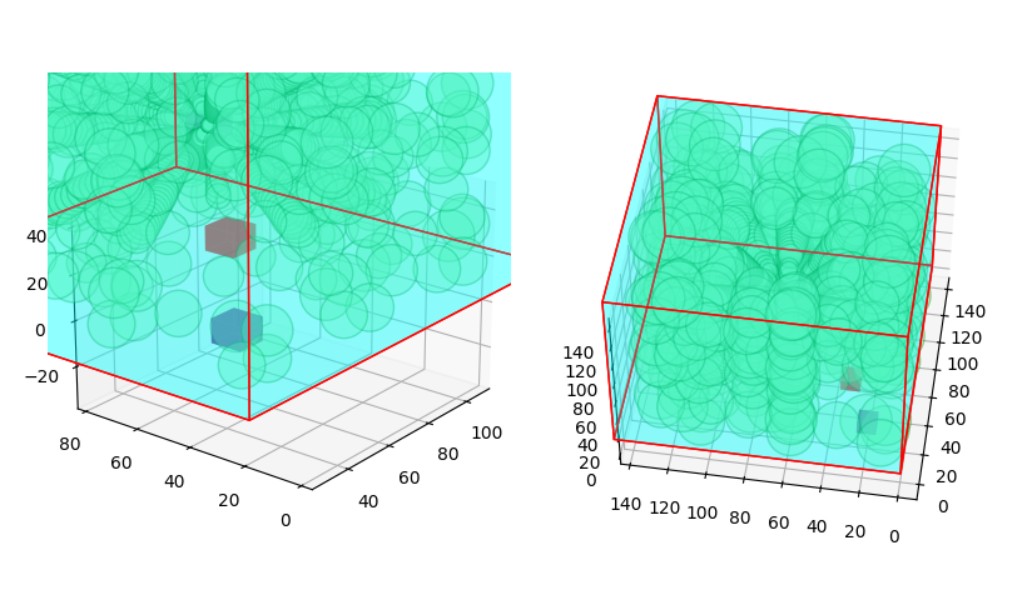}
\end{center}
\begin{center}
  \caption{The Rough Mereological Potential Field algorithm constructed in 3-dimensional environment. The below case represents the goal point located at the point (120,120,120) represented as a blue cube with three red obstacles. The bigger cube represents the gate for a robot. }\label{square_algorithm3}
  \includegraphics[scale=0.35]{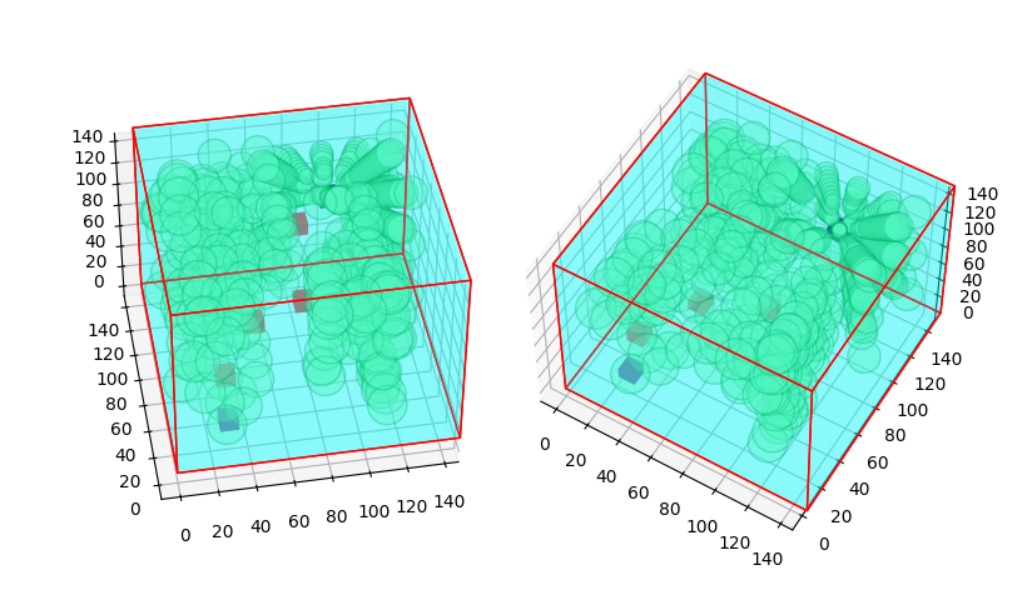}
\end{center}
\end{figure}

The idea of an algorithm is to generate the neighbors around the main point (goal). In the 2D concept, there were eight crucial neighbors, which determined the potential field position in each iteration. The 3D case became more complex. Adding the third dimension 'z' forced us to implement more neighbors to be considered in the potential field generating. We have to consider the constant value 'z', increasing 'z' and decreasing 'z'. The mentioned operation gave us 24 neighbors, instead of 8 (N, W, S, E, NW, NE, SW, SE) which were in the 2-dimensional space. The number of neighbors depends on directions. Below we can see the main assumptions of the structure of the algorithm and also the result that we obtained - Figs. \ref{square_algorithm}, \ref{square_algorithm2} and \ref{square_algorithm3}.

\newpage
\begin{enumerate}
\item Declare the basic values:
\begin{center}
- Set the current distance to the goal:
d = 0,\newline 
- Put the direction value on standby 
clockwise = True\newline
\end{center}
\item Prepare an empty queue $Q$:
\begin{center}
     $Q = \emptyset$
\end{center}

\item Add into created queue $Q$ first potential field p(x,y,z,d), where $x,y,z$ express the location coordinates of already created field and $d$ reflects the current distance to the goal:

\begin{center}
    $Q \cup \{p(x,y,z,d)\}$
\end{center}

\item Iterate through $Q$,
\begin{enumerate}

\item Determine neighbors created with respect to a current direction:\newline

if clockwise is true:

\begin{center}
\begin{small}
   $$
N = \left\{ \begin{array}{ll}
p_0 = p(x-d,y,z,d),\\
p_1 = p(x-d,y+d,z,d),\\
p_2 = p(x,y+d,z,d),\\
p_3 = p(x+d,y+d,z,d),\\
p_4 = p(x+d,y,z,d),\\
p_5 = p(x+d,y-d,z,d),\\
p_6 = p(x,y-d,z,d),\\
p_7 = p(x-d,y-d,z,d)\\
p_8 = p(x-d,y,z-d,d),\\
p_9 = p(x-d,y+d,z-dd),\\
p_{10} = p(x,y+d,z-d,d),\\
p_{11} = p(x+d,y+d,z-d,d),\\
p_{12} = p(x+d,y,z-d,d),\\
p_{13} = p(x+d,y-d,z-d,d),\\
p_{14} = p(x,y-d,z-d,d),\\
p_{15} = p(x-d,y-d,z-d,d)\\
p_{16} = p(x-d,y,z+d,d),\\
p_{17} = p(x-d,y+d,z+d,d),\\
p_{18} = p(x,y+d,z+d,d),\\
p_{19} = p(x+d,y+d,z+d,d),\\
p_{20} = p(x+d,y,z+d,d),\\
p_{21} = p(x+d,y-d,z+d,d),\\
p_{22} = p(x,y-d,z+d,d),\\
p_{23} = p(x-d,y-d,z+d,d)\\
\end{array} \right\}
$$
\end{small}
\end{center}
\newpage
if anticlockwise is true:
\begin{center}
\begin{small}
  $$
N' = \left\{ \begin{array}{ll}
p_0 = p(x-d,y-d,z,d),\\
p_1 = p(x,y-d,z,d),\\
p_2 = p(x+d,y-d,z,d),\\
p_3 = p(x+d,y,z,d),\\
p_4 = p(x+d,y+d,z,d),\\
p_5 = p(x,y+d,z,d),\\
p_6 = p(x-d,y+d,z,d),\\
p_7 = p(x-d,y,z,d)\\
p_8 = p(x-d,y-d,z-d,d),\\
p_9 = p(x,y-d,z-d,d),\\
p_{10} = p(x+d,y-d,z-d,d),\\
p_{11} = p(x+d,y,z-d,d),\\
p_{12} = p(x+d,y+d,z-d,d),\\
p_{13} = p(x,y+d,z-d,d),\\
p_{14} = p(x-d,y+d,z-d,d),\\
p_{15} = p(x-d,y,z-d,d)\\
p_{16} = p(x-d,y-d,z+d,d),\\
p_{17} = p(x,y-d,z+d,d),\\
p_{18} = p(x+d,y-d,z+d,d),\\
p_{19} = p(x+d,y,z+d,d),\\
p_{20} = p(x+d,y+d,z+d,d),\\
p_{21} = p(x,y+d,z+d,d),\\
p_{22} = p(x-d,y+d,z+d,d),\\
p_{23} = p(x-d,y,z+d,d)\\
\end{array} \right\}
$$
\end{small}
\end{center}
\item Count the Euclidean distance from neighbors created in the previous iteration and already generated potential fields  to avoid the superfluous fields,
\item If  Euclidean distance of topical potential field $p_k(x,y,z,d)$  is less than 15 and ${p_k(x,y,z,d)\cup O}$, where $O$ is the collection of obstacles coordinates, what's more, ${p_k(z,y,d)\cup F}$, where $F$ comprising potential fields created to meet the aforementioned conditions, then the current field is to be abandoned. In such a scenario, the system is required to backtrack to point 4,\newline
\item Find out if in $Q$ is any identical potential field if exists drop the current neighbor, and go back to point 4,\newline
\item After the end of filtration add the accepted potential field $p_k(x,y,z,d)$ into the end of list $Q$,\newline
\item Expand the distance value to the goal:
\begin{center}
    $d = d(p_k)+0.5$
\end{center}
\item Change the direction to negative,\newline
if clockwise = True:\\
     change direction into anticlockwise = True, clockwise = False\newline
if anticlockwise = True:\\
    change direction into clockwise = True, anticlockwise = False
\item Drop current neighbour $p_k(x,y,z,d)$ from dynamic queue $Q$ and attach it into potential fields list $F$.
\end{enumerate}
\end{enumerate}

According to the presented way, the distance is initialized from value {0}. This number will be increasing in the next iterations because our algorithm starting to generate potential fields from a goal into a starting point. Created potential fields are represented by $p(x,y,z,d)$, where $x,y,z$ are proper coordinates and $d$ describes the value of a distance, which during the operation is responsible for generating new neighbors. The bigger the distance is, the further the potential field has been created from a goal. Moreover in each iteration, the direction of generating neighbors has to be changed so as not to get stuck and explore all map. 
\newline
Below and in the next sections we are going to show the visualization of the results. The term 'bigger cube' describes the cube, which is almost equal to the scale of our plot. The term 'smaller cube' determines the figure, which is inside the 3D plot, and hence it is inside the 'bigger cube'. 

\clearpage

\clearpage

\subsection{Path finding}
To generate the path, we are using the variation of the algorithm proposed by Osmialowski Path Search Algorithm \cite{Osmialowski 2011} and later modifying by Szpakowska and Artiemjew \cite{SzpakowskaArtiemjew 2023}. Above mentioned algorithm within the given potential field is as follows:

\begin{enumerate}
    \item Start iterating the loop from a list of fields given from the Adapted Square fill algorithm \ref{sqr_alg}.
    \item Take the first generated potential field from the list of fields (generated as the last one), this is the \textit{actual field}.
    \item Count the Weighted Euclidean Distance between the \textit{actual field}, the next potential field in the list of fields - \textit{next field}, and the goal.
    \item Look for the minimum counted distance and remember the \textit{next field} - that field which gave the minimum distance will be the \textit{actual field} in the next iteration. To accept the distance and the field it is necessary to check if the connection between the actual field and the chosen field is possible. That means we have to take into account if the path between them does not cross any obstacle. 
    \item If our loop finishes for a given current potential field we have to change the \textit{actual field} to the chosen field with the smallest Weighted Euclidean Distance. 
    \item The operation is going to be reaped for the new \textit{actual field}. The algorithm stops when the \textit{actual field} is close or equal to the goal point. 
\end{enumerate}

\begin{figure}[ht]
\begin{center}
  \caption{Generated path using the Path Search Algorithm. The darkened cubes determine the start and goal. Lighter cubes are obstacles. }
  \includegraphics[scale=0.45]{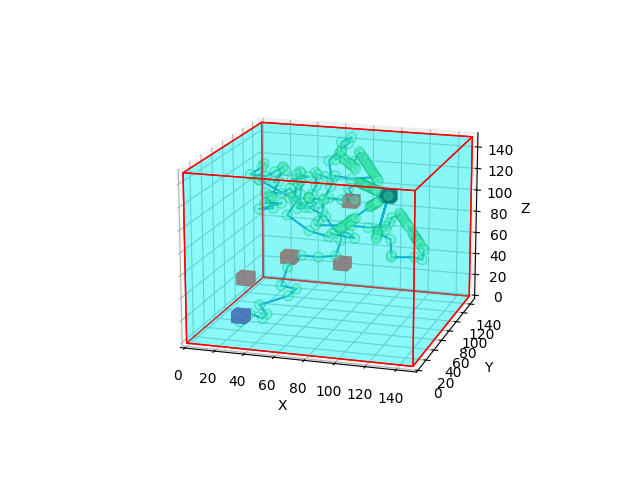}\label{square_first_path}
\end{center}
\end{figure}
\clearpage
\subsubsection{Distance counting - weighted Euclidean distance}\label{subsection : distance counting}

To determine the distance between two potential fields we used classical Euclidean distance and variation of this distance:

\begin{center}
$d(p,q) = \sqrt{\sum_{i=1}^{n} (p_i - q_i)^2}$
\end{center}
In the context of a three-dimensional Euclidean plane, wherein $p$ denotes the coordinates of a point characterized by potential fields, and $q$ signifies the point with goal coordinates, an enhancement to path-finding accuracy is introduced by incorporating two distinct Euclidean distances. Specifically, the first distance, referred to as the classical Euclidean distance in this study, focuses on measuring the distance between the current potential fields. The second distance, denoted as the Weighted Euclidean distance, is utilized for searching the first basic path, and the classic one is used for optimization.\\

The Weighted Euclidean distance accounts for both the distance between the goal and the current point. Additionally, this distance is subjected to a weight, achieved by multiplying it with a specific floating-point value. The weight is applied to both the distance between the goal and the current point and the distance between two potential fields. This utilization of weighted distance aims to mitigate the risk of selecting a sub-optimal path, such as a direct leap to the target without due consideration for obstacles.

\subsection{Field filtering- path optimizing}\label{subsection : path optimizing}
The resultant path obtained from the Path Search Algorithm did not meet our anticipated standards of optimality and clarity. To diminish the noise within the path, we implemented a path optimization filter. This filter centers its attention on the distances between points in the generated path and the designated goal.\\

The primary condition stipulates the following: commencing from the initial element in the path if the distances from the current points to the target are either identical or greater, those particular points are excluded. In instances where multiple points exhibit the same distance, a secondary step is introduced. We calculate the distances from the neighboring points to the target and compare them. The points characterized by the smallest values of distance are retained, ensuring a refined and more optimal path.\\
\clearpage

\begin{figure}[ht]
  \centering
  \caption{The same path after applied filtering for optimization. Not needed potential fields were truncated, in the result we get a clear path. }\label{square_optimal_path}
    \includegraphics[scale=0.7]{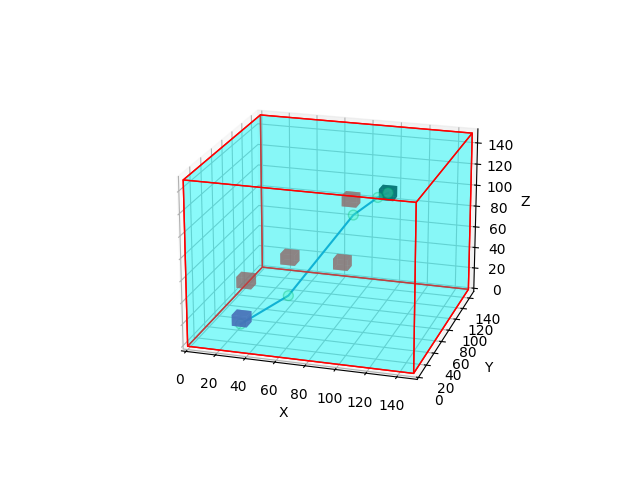}
    
  \hspace{1cm} % Adjust the horizontal space between subfigures
  
  \caption{Another view of the same path after applied filtering for optimization.}
  \label{square_optimal_path2}
    \includegraphics[scale=0.7]{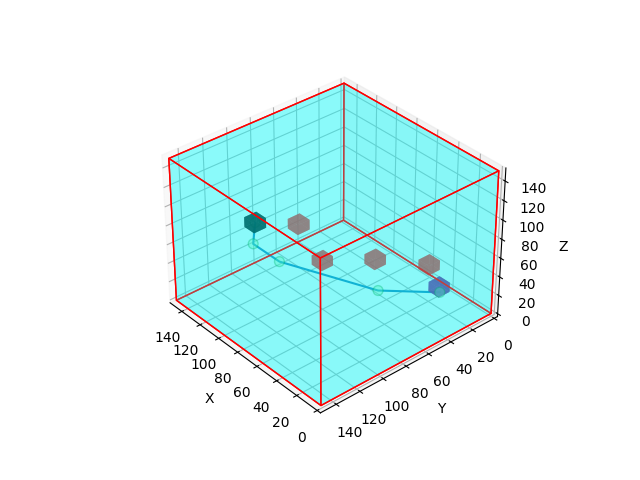}
\end{figure}

\clearpage

\subsection{Path smoothing}\label{Path smoothing}
 Upon visualizing the optimal path from the robot's starting location to the target, we proceed to initialize the path smoothing algorithm introduced by Zmudzinski and Artiemjew in 2017 \cite{Zmudzinski Artiemjew 2017}. The algorithm is iteratively applied '$n$ times until the obtained result and the shape of the path meet the desired satisfaction:

 \begin{enumerate}
     \item We minimize the distance between points by incorporating the variable $\alpha$, which dictates the rate at which we deviate from the original position  $x_k$ its adjustment considers both the preceding point $x_k-1$ and the subsequent point $x_k+1$ influencing the movement away from the initial position.
     \begin{center}
    $x_k = x_k +\alpha(x_k-1+x_k+1-2x_k)$\\
    $y_k = y_k +\alpha(y_k-1+y_k+1-2y_k)$\\
    $z_k = z_k +\alpha(z_k-1+z_k+1-2z_k)$
     \end{center}
     \item Subsequently, we introduce a balancing step for the point $x_k$ through the application of the $\beta$ variable. This entails calculating $y_k$ which signifies the new position of the point. By performing this operation, we aim to mitigate the tendency towards a straight-line trajectory in the path.
     \begin{center}
         $x_k = x_k +\beta(x_k-x_k)$\\
         $y_k = y_k +\beta(y_k-y_k)$\\
         $z_k = z_k +\beta(z_k-z_k)$
     \end{center}
 \end{enumerate}

\begin{figure}
    \centering
    \caption{Comparing of path creation. From left: first basic path, filtered path, smoothed path.}\label{combine3}
    \includegraphics[scale=0.2]{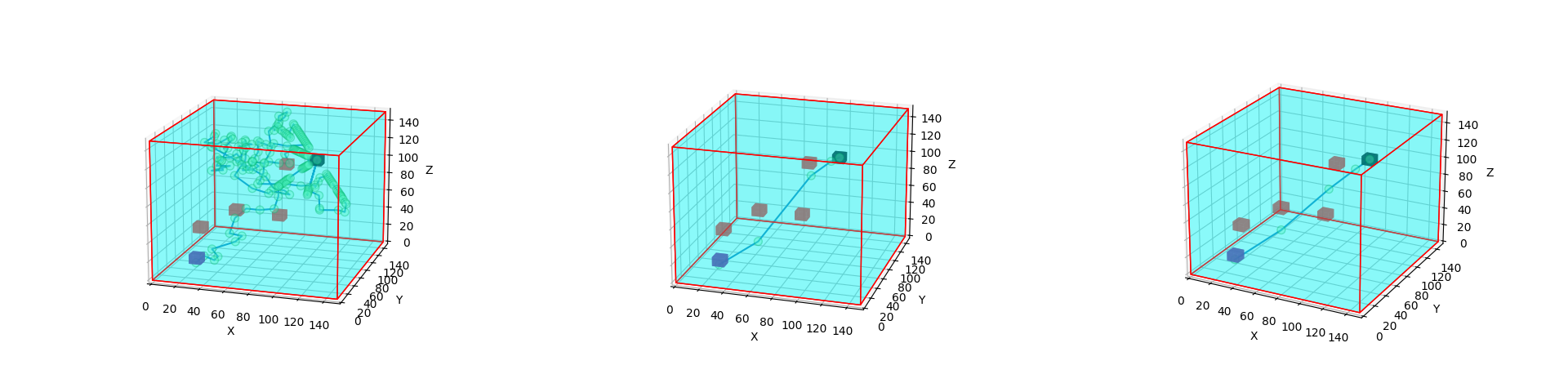}

\end{figure}

The results are shown in figure \ref{combine3}. The comparison of all paths shows that we eliminated unnecessary points in the path, and as a result, our path became more unambiguous. After applying the path smoothing algorithm we get a possibility to achieve a better trajectory of a future robot movement. 

% \clearpage
% \begin{figure}[ht]

% \begin{center}
%   \caption{The same path after applied filtering for optimization. The solution below provides us with a smooth path, without sharp edges.}\label{smoothed_path1}
%   \includegraphics[scale=0.7]{img/smooth_3.png}
% \end{center}
% \end{figure}
\section{Experimential section} \label{experiments}

This section will concentrate on the underlying technical components. Initially, we will outline the programming language and associated libraries responsible for various facets of the project. Subsequently, we will explore the preparatory steps for the environment and the requisite conditions that need to be established. 
\newline
It is worth mentioning that determining the boundaries of the cage and the positioning of obstacles, start, and goal are flexible. The fact that the created virtual world returns correct information about the location of given points depends largely on the correctly declared real distance between the frame boundaries (frame height and width). The localization of declared points is in real-time, so there is a possibility to follow the actual robot position and the location of obstacles. In that case, we can create different maps, with another number and combination of obstacles, goal, and start - actual point.
\subsection{Technical aspects} \label{technical aspects}

The entire project was developed using Python \cite{Python}. The key libraries crucial for this paper include Matplotlib and OpenCV. Specifically, the Poly3DCollection package \cite{Matplotlib} was utilized for 3-dimensional visualization. Additionally, the OpenCV library \cite{OpenCV} played a pivotal role in the creation of the project environment.

\subsection{Environment preparation} \label{environment}
The environment had been created using Python library \cite{OpenCV}. Two cameras were hung in space, one on the top of a gate and another one aside. Two types of color markers were placed, which determine the borders of a gate. The length between markers was measured, so all of the coordinates of points inside the gate were read concerning the created world dimensions. \newline
The initial phase of the preparatory process involved coding an algorithm for color recognition. Subsequently, color markers were strategically positioned within the laboratory space, and their sequence determined the order for color reading. This procedural step enabled the program to identify the recognized wall, with each camera assigned to recognize a specific color sequence. Following this, the read values were converted into real-world measurements by assessing the distance between designated markers, which were then transposed into the virtual world. This operation facilitated the extraction of real-world coordinates for various points within the environment, including the goal point, starting point, and obstacle points.
\begin{figure}
    \centering
    \caption{Visualization of a generated gate by one-painted color markers.
    This example of environment creation is visualized in Fig.\ref{square_algorithm}, where there was only a goal point inside the cadge. In that visualization, we show the idea of naming the 'bigger cube' and the 'small cube'. The bigger cube is a virtual cadge determined by using color markers, the smaller cube is a solid in space.}
    \includegraphics[scale=0.3]{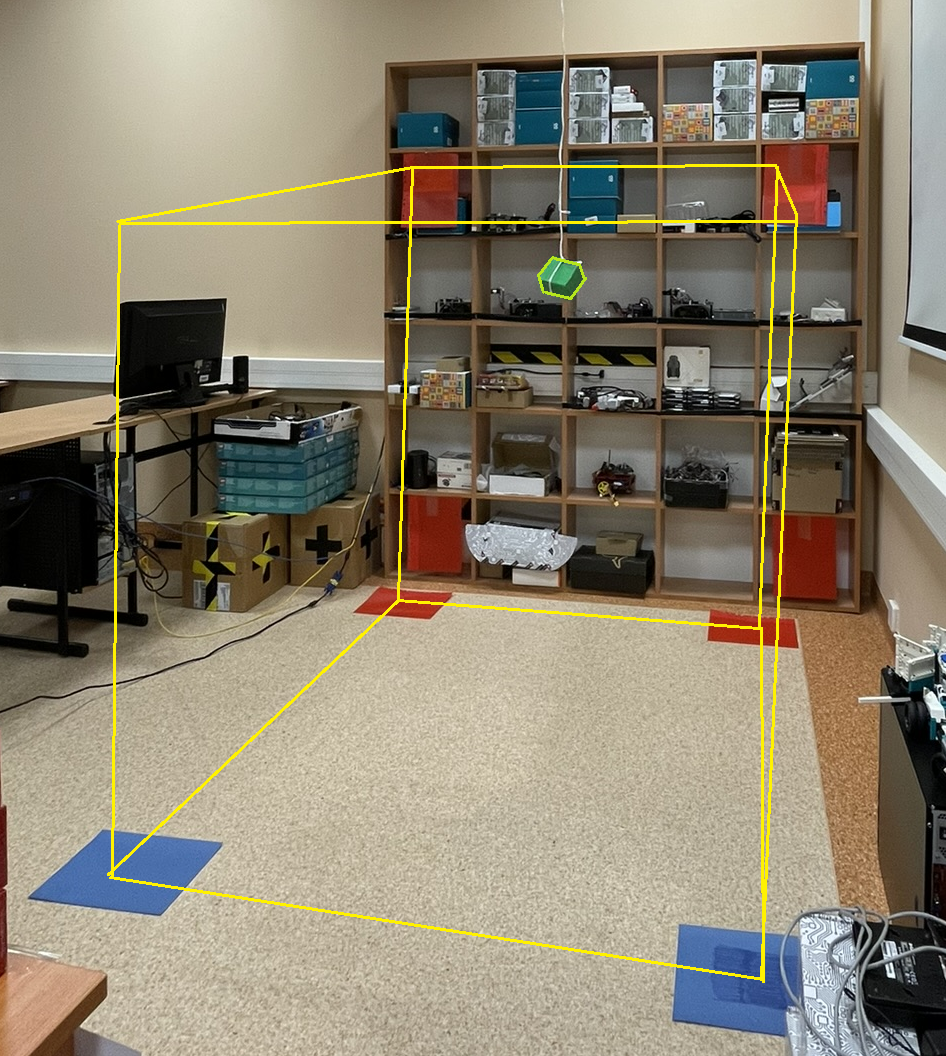}
    
    \label{fig:environment}
\end{figure}

The visualization of the prepared environment is depicted in figure \ref{fig:environment}. Within the virtual gateway, only the goal point was positioned—a single colored cube suspended in space. This gateway delineates the designated space for drone movement.
% \clearpage

The complete set of codes for this project is accessible on \cite{github:project}. To replicate the project, access to two cameras functioning in the real world is essential, along with the installation of the required libraries. Additionally, recreating the gate necessitates arranging colored markers in the specified configuration.

\section{Conclusions}\label{conclusions}
In this investigation, we effectively expanded and implemented a path planning algorithm within a three-dimensional environment, employing a rough mereological potential field. Furthermore, we conducted path optimization and subsequent smoothing, focusing on three dimensions. Real-time visualization reflecting the actual environmental conditions was achieved through the utilization of images from two cameras. The objective of the study was successfully realized, establishing the conditions and framework for the optimal navigation of a moving machine in three-dimensional real-time, particularly on a map with obstacles. This accomplishment represents a significant stride toward the practical application of the rough mereological potential field in drone-related scenarios. 
\newline

\section{Acknowledgements}
This work has been supported by a grant from the Ministry of Science and Higher Education of the Republic of Poland under project number 23.610.007-110

% \bibliographystyle{splncs04}
% \bibliography{bibliografia}

\begin{thebibliography}{99}

\bibitem{OpenCV}
OpenCV, \url{https://opencv.org/}

\bibitem{polkowski2008} Polkowski, L., Osmialowski, P.: \emph{A Framework for Multiagent Mobile Robotics: Spatial Reasoning Based on Rough Mereology in Player/Stage System.}, pages: 142-149 (2008)

\bibitem{Osmialowski2008} Osmialowski, P. (2013). On path planning for mobile robots: introducing the mereological potential field method in the framework of mereological spatial reasoning. Journal of Automation, Mobile Robotics and Intelligent Systems, 3(2), 24-33.

% \bibitem{OsmialowskiPolkowski 2010} Osmialowski, P., Polkowski L.: \emph{Spatial Reasoning Based on Rough Mereology: A Notion of a Robot Formation and Path Planning Problem for Formations of Mobile Autonomous Robots.} LNCS. Trans. Rough Sets 12: 143-169 (2010)

\bibitem{Polkowski 1996}
Polkowski, L.: \emph{Rough Mereology: A new paradigm for approximate reasoning,} In: International Journal of Approximate Reasoning, (1996)

\bibitem {SzpakowskaArtiemjew 2023} Szpakowska, A., Artiemjew, P., Cybowski, W. \emph{Navigational Strategies for Mobile Robots Using Rough Mereological Potential Fields and Weighted Distance to Goal.} Rough Sets. IJCRS 2023., Springer, vol 14481, pages 549-564, (2023)

\bibitem{Osmialowski 2011} Osmialowski, P.: (2022). \emph{Planning and Navigation for Mobile Autonomous Robots Spatial Reasoning in Player/Stage System} (2011)

\bibitem{Polkowski 2018}
Polkowski, L., Zmudzinski, L., Artiemjew, P.: \emph{Robot navigation and path planning by means of rough mereology, }Proceeding of the IEEE International Conference on Robotic Computing, pages 363-368 (2018)

\bibitem{Zmudzinski Artiemjew 2017} Zmudzinski L., Artiemjew, P.: \emph{Path planning based on potential fields from rough mereology,} Rough Sets. International Joint Conference, IJCRS 2017, Olsztyn, Poland, pages 158-168 (2017)

\bibitem{Python} \url{https://www.python.org/}

\bibitem{Matplotlib} \url{https://matplotlib.org/3.1.1/api/_as_gen/mpl_toolkits.mplot3d.art3d.Poly3DCollection.}

\bibitem{github:project}
Github Szpakowska, \url{https://github.com/aleksandraszpakowska/3D_Rough_Mereology}



\end{thebibliography}

\end{document}